\pdfoutput=1

\documentclass[11pt]{article}

\usepackage{acl}

\usepackage{times}
\usepackage{latexsym}

\usepackage{microtype}

\usepackage{url}
\usepackage{tipa}

\usepackage{amsmath}

\usepackage{caption}
\usepackage{subcaption}

\usepackage{todonotes}

\usepackage{soul}
\makeatletter
\newcommand*\iftodonotes{\if@todonotes@disabled\expandafter\@secondoftwo\else\expandafter\@firstoftwo\fi}  
\makeatother

\newcommand{\citeposs}[1]{\citeauthor{#1}'s (\citeyear{#1})}

\usepackage{cleveref}
\crefname{section}{\S}{\S\S}
\Crefname{section}{\S}{\S\S}
\crefname{table}{Table}{}
\crefname{figure}{Figure}{Figs.}
\crefname{algorithm}{Algorithm}{}
\crefname{algorithm}{Algorithm}{}
\crefname{line}{Line}{}
\crefname{appendix}{Appendix}{}
\crefformat{section}{\S#2#1#3}
\crefname{thm}{Theorem}{}
\crefname{cor}{Corollary}{}
\crefname{prop}{Proposition}{}
\crefname{def}{Definition}{}

\usepackage[T1]{fontenc}

\usepackage[utf8]{inputenc}

\usepackage{linguex}

\usepackage{graphicx}
    
\title{Subject Verb Agreement Error Patterns in Meaningless Sentences: Humans vs. BERT}

\newcommand{\lattice}{\normalfont \text{\textipa{@}}}
\newcommand{\unipi}{\normalfont \text{\textipa{B}}}

\author{
Karim Lasri$^{\lattice,\unipi}$ Olga Seminck$^{\lattice}$ Alessandro Lenci$^{\unipi}$ Thierry Poibeau$^{\lattice}$ \\
$^{\lattice}$Lattice (\'Ecole Normale Supérieure-PSL, CNRS, U. Sorbonne Nouvelle)~\; \\ 
~$^{\unipi}$University of Pisa~\;  \\
  \texttt{\href{mailto:karim.lasri@ens.psl.eu}{karim.lasri@ens.psl.eu}}~\;~ 
  \texttt{\href{mailto:olga.seminck@cnrs.fr}{olga.seminck@cnrs.fr}}~\;~ \\
  \texttt{\href{mailto:alessandro.lenci@unipi.it}{alessandro.lenci@unipi.it}}~\;~ \texttt{\href{mailto:thierry.poibeau@ens.psl.eu}{thierry.poibeau@ens.psl.eu}}
}

\date{}

\begin{document}
\maketitle
\begin{abstract}
Both humans and neural language models are able to perform subject-verb number agreement (SVA). In principle, semantics shouldn't interfere with this task, which only requires syntactic knowledge.
In this work we test whether meaning interferes with this type of agreement in English in syntactic structures of various complexities. To do so, we generate both semantically well-formed and nonsensical items. We compare the performance of BERT-base to that of humans, obtained with a psycholinguistic online crowdsourcing experiment. We find that BERT and humans are both sensitive to our semantic manipulation: They fail more often when presented with nonsensical items, especially when their syntactic structure features an attractor (a noun phrase between the subject and the verb that has not the same number as the subject). We also find that the effect of meaningfulness on SVA errors is stronger for BERT than for humans, showing higher lexical sensitivity of the former on this task.
\end{abstract}


\section{Introduction}

Subject Verb Agreement (SVA) is a grammatical constraint in English and several other natural languages, such that verbs must agree in number with their subject. Linguistic theories generally assume that SVA obeys two main principles: i.) \textbf{structure dependence} (SD) - SVA is governed by phrasal structure, rather than surface linear order (i.e., the verb agrees with the syntactic subject);
 ii.) \textbf{meaning independence} (MI) - SVA is a morphosyntactic constraint that holds for meaningless sentences too (e.g., \textit{Colorless green ideas sleep furiously}) \citep{chomsky1956three,chomsky1971problems,chomsky1976reflections}. 
 However, previous research has shown that humans are prone to making agreement errors with specific constructions \cite{bock1991broken,hartsuiker2001object}, for example when a noun with a different number (also called an \textbf{attractor}) occurs between the subject (the \textbf{cue}) and the verb (the \textbf{target}).
See~\ref{ex_bock_miller} from \citet{bock1991broken} for an example where agreement can be disturbed by an attractor, as human subjects often show preference for a syntactically ill-formed sentence:

\ex. [The \textbf{readiness}]\textsubscript{subject} [of our conventional \underline{forces}\textsubscript{attractor}]\textsubscript{PP} [\textit{are}]\textsubscript{verb} at an all-time low.\label{ex_bock_miller} 

This evidence suggests that the SD principle of SVA  might be weaker than it is typically assumed and can be disrupted or disturbed under specific conditions. At the same time, such violation prompts the need to carefully test whether the MI principle of SVA is also compromised in subjects' grammaticality judgments. 

SVA has become a widespread testbed to investigate the syntactic knowledge that \textbf{neural language models} (NLMs) are able to acquire. A key question is to explore to what extent their competence of syntax obeys the same constraints as those of humans, by comparing the behavior of NLMs with subjects' judgments. In this paper, we pursue this goal by focusing on the SD and MI properties of SVA in humans and NLMs. Its contribution is twofold: i.) we collect original human data on meaningful and meaningless stimuli featuring syntactic structures of varying complexities; ii.) we analyze and compare the error patterns in humans and in NLMs. This allows us to address the following questions: i.) are the SD and MI assumptions about SVA truly supported by human judgments? ii.) do humans and NLMs make similar SVA errors in structures with attractors and/or in meaningless sentences? This may help understand to what extent NLMs rely on syntactic knowledge abstracted from training examples. By comparing the error patterns of humans and NLMs on SVA, we can derive important information about the the nature of their linguistic competence: Is the ability of NLMs on SVA completely meaning independent? Is it influenced by the complexity of the syntactic structure?


\begin{table*}
\begin{small}
\begin{center}
\begin{tabular}{ cll } 
 \hline
 \textbf{Structure} & \textbf{Structure description} & \textbf{Example} \\
 \hline
 \textbf{A} & Simple agreement & [The \textbf{author}]\textsubscript{subject} [\textit{laughs/*laugh}]\textsubscript{verb} \\
 \textbf{B} & In a sentential complement & [The mechanics]\textsubscript{subject} [said]\textsubscript{VComp} [[the \textbf{author}]\textsubscript{SComp} [\textit{laughs/*laugh}]\textsubscript{verb}]\textsubscript{Comp} \\
 \textbf{C} & Across a prepositional phrase & [The \textbf{mechanic}]\textsubscript{subject} [near the \underline{author}]\textsubscript{PP} [\textit{smiles/*smile}]\textsubscript{verb} \\
 \textbf{D} & Across a subject relative clause & [The \textbf{author}]\textsubscript{subject} [that likes the \underline{movie}]\textsubscript{subj. RC} [\textit{laughs/*laugh}]\textsubscript{verb} \\
 \textbf{E} & In a short verb phrase coordination & [The \textbf{author}]\textsubscript{subject} [laughs and \textit{swims/*swim}]\textsubscript{VP Coord} \\
 \textbf{F} & Across an object relative clause & [The \textbf{author}]\textsubscript{subject} [that the \underline{mechanics} like]\textsubscript{obj. RC} [\textit{smiles/*smile}]\textsubscript{verb} \\
 \textbf{G} & Within an object relative clause & [The author]\textsubscript{subject} [that the \textbf{mechanics} \textit{like/*likes}]\textsubscript{obj. RC} [smiles]\textsubscript{verb} \\
\textbf{H} & Across an object relative clause (\textit{no that}) & [The \textbf{author}]\textsubscript{subject} [the \underline{mechanics} like]\textsubscript{obj. RC} [\textit{smiles/*smile}]\textsubscript{verb} \\
 \textbf{I} & Within an object relative clause (\textit{no that}) & [The author]\textsubscript{subject} [the \textbf{mechanics} \textit{like/*likes}]\textsubscript{obj. RC} [smiles]\textsubscript{verb} \\
 \hline
\end{tabular}
\caption{\label{tab:agreement-structures}Agreement structures used in this study, from \citet{marvin-linzen-2018-targeted}. The cue is bolded, the target is in italic, and the attractor is underlined. For each target, we display both the correct and incorrect inflections.}
\end{center}
\end{small}
\end{table*}

\section{Related Work}

\citet{linzen-etal-2016-assessing} tested Long Short Term Memory (LSTM) language models and their performance on the SVA task. They found that these models can capture a non-trivial amount of grammatical structure but that they are insufficient for capturing complex syntax-sensitive dependencies. 
In \citet{marvin-linzen-2018-targeted} the capacity of LSTM language models to perform the SVA task is compared with human  data. In particular for sentences including attractors, the models perform worse than humans. \citet{linzen_leonard} tested recurrent neural networks (RNNs) on the task to see whether the error pattern was similar to humans. They concluded that despite important similarities, there was a different behavior within relative clauses. In particular, RNNs are sensitive to the number of attractors, whereas humans are not.

\citet{gulordava-etal-2018-colorless} investigated the role of semantics on the performance of RNNs on SVA, testing such neurals models against meaningless, or ``nonce'', sentences built from various syntactic constructions.
They used meaningless sentences where RNNs could not rely on semantic or lexical cues. For Italian, they found that the performance of an LSTM model and the performance of humans were comparable for meaningless items and semantically congruent ones. \citet{lasri-etal-2022-bert} tested a transformer model, BERT-base \cite{devlin-etal-2019-bert}, on its capacity of subject-verb agreement on the items from \citet{marvin-linzen-2018-targeted} and on semantically disrupted sentences featuring the same syntactic constructions. They found that although the model generalized well for simple templates, it failed on meaningless items with attractors.

\section{Experimental Setup}

In this section, we describe the procedure to construct the experimental items used to collect human judgments with crowdsourcing and to test NLM behavior on SVA.

\paragraph{Items}
We test the syntactic structures in \cref{tab:agreement-structures}, four of which include an attractor. For each syntactic template, we generate $30$ meaningless sentences by sampling random words with the correct part-of-speech at each position of the template, using an extensive vocabulary (see \cref{selection_voc} for more details). The generated sentences are meaningless in the sense that our generation procedure does not include any semantic constraint designed to grant meaning to the sentence. The resulting word sequence should therefore be semantically ill-formed, similar to the classic example \emph{Colorless green ideas sleep furiously}. As some generated sentences can accidentally happen to be interpretable, we also tried to manually remove such sentences when crafting our stimuli. However, as meaning is a subtle notion, some sentences might be interpretable in very specific contexts after our manual filtering. Still, we expect such cases to be very rare in our dataset. 
Every minimal pair consists of sentences similar to ~\ref{admin_sings}. 

\ex. \label{admin_sings} \a. *The admissions sings.\label{ex_a} 
\b. The admissions sing. \label{ex_b} 

We also sample 30 meaningful sentences for each structure from \citet{marvin-linzen-2018-targeted} to collect human data.\footnote{We will call this data \textbf{M\&L}. We filter out sentences where the verb is `be', as this verb is very frequent in English.}
As the filtered M\&L data used to collect human performances resulted in a very limited number of sentences built on a limited vocabulary, we extract BERT's performance on the whole dataset for M\&L, following the procedure in \citet{goldberg}.

We thus collect human performance on 30 items for each of our 2 conditions (Nonce and M\&L), and each of our 9 syntactic structures, for a total of 540 items. More details are given in \cref{app:collection_hum_data}.\footnote{Our data is available at \url{https://github.com/karimlasri/agreement-humans-bert}}

\paragraph{Neural Language Model}
Our tested NLM is BERT \cite{devlin-etal-2019-bert}, a bi-directional pre-trained transformer model. BERT has been shown to possess a number of syntactic abilities \cite{jawahar-etal-2019-bert, goldberg}, the nature of which remain scarcely understood \cite{baroni-generalization}. 
For each item, we present BERT with sentences masked at the target position and compare the probability that BERT assigns to each verb inflection. 

\begin{figure}
    \begin{center} \includegraphics[width=\columnwidth]{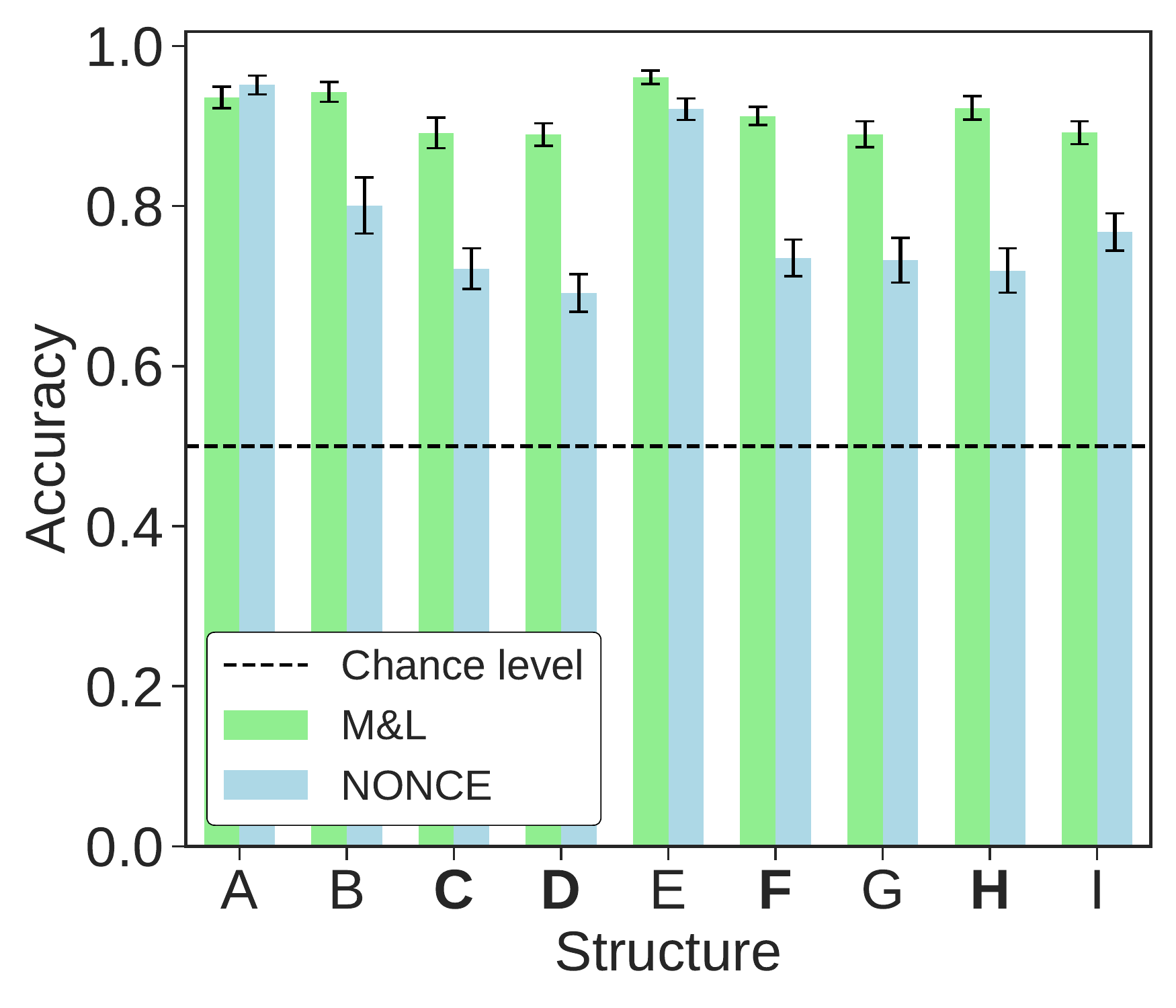}
    \caption{Human accuracies on the SVA task. Structures where an attractor is present are displayed in bold. The error bars displayed represent the 95\% confidence interval.}
    \label{fig:raw-accuracies}
    \end{center}
\end{figure}

\begin{figure*}
    \centering
    \begin{subfigure}{0.45\textwidth}
        \includegraphics[width=\textwidth]{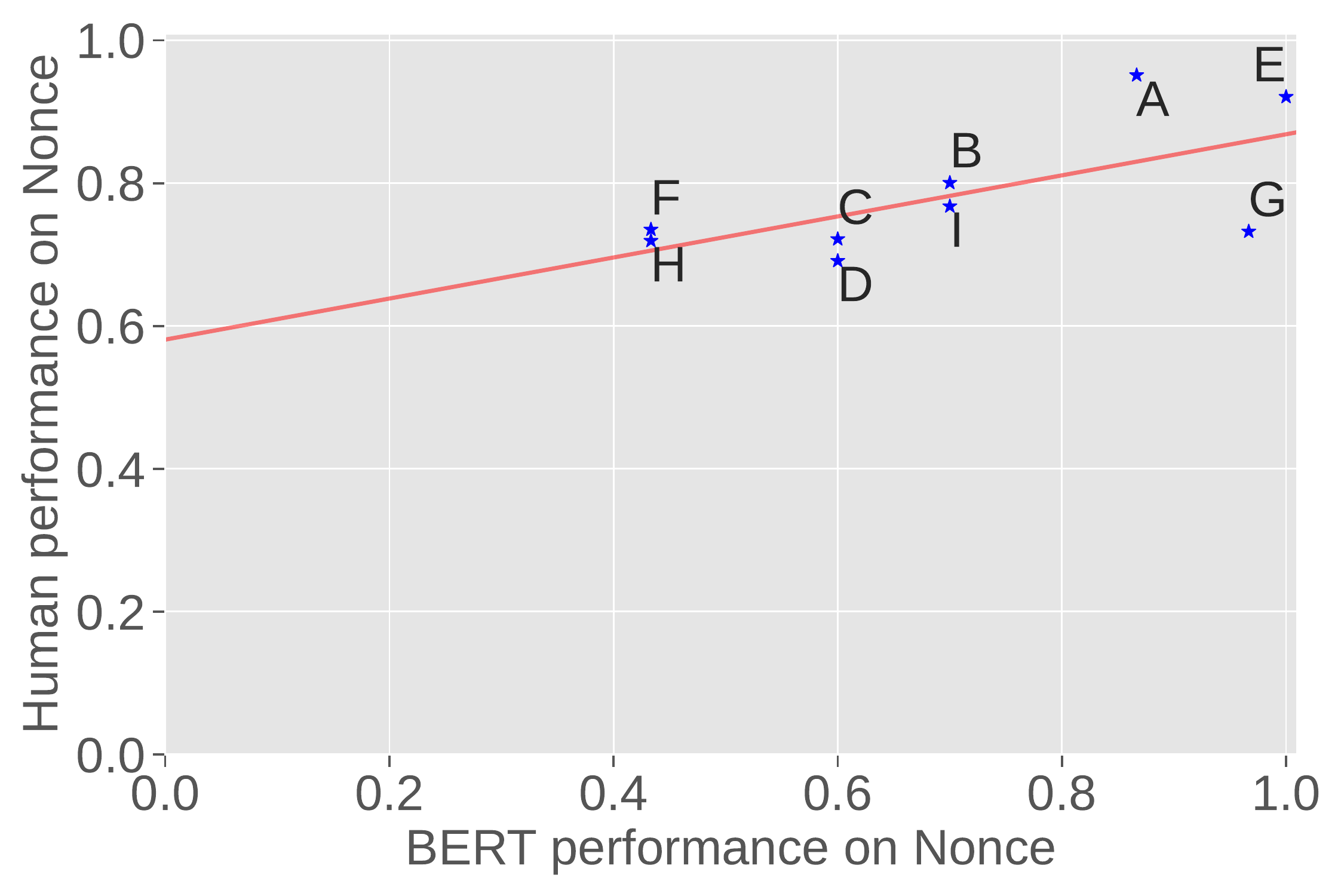}
        \subcaption{Human nonce performance vs. BERT}
        \label{fig:nonce-bert-humans}
    \end{subfigure}
    \hspace{5mm}
    \begin{subfigure}{0.45\textwidth}
        \includegraphics[width=\textwidth]{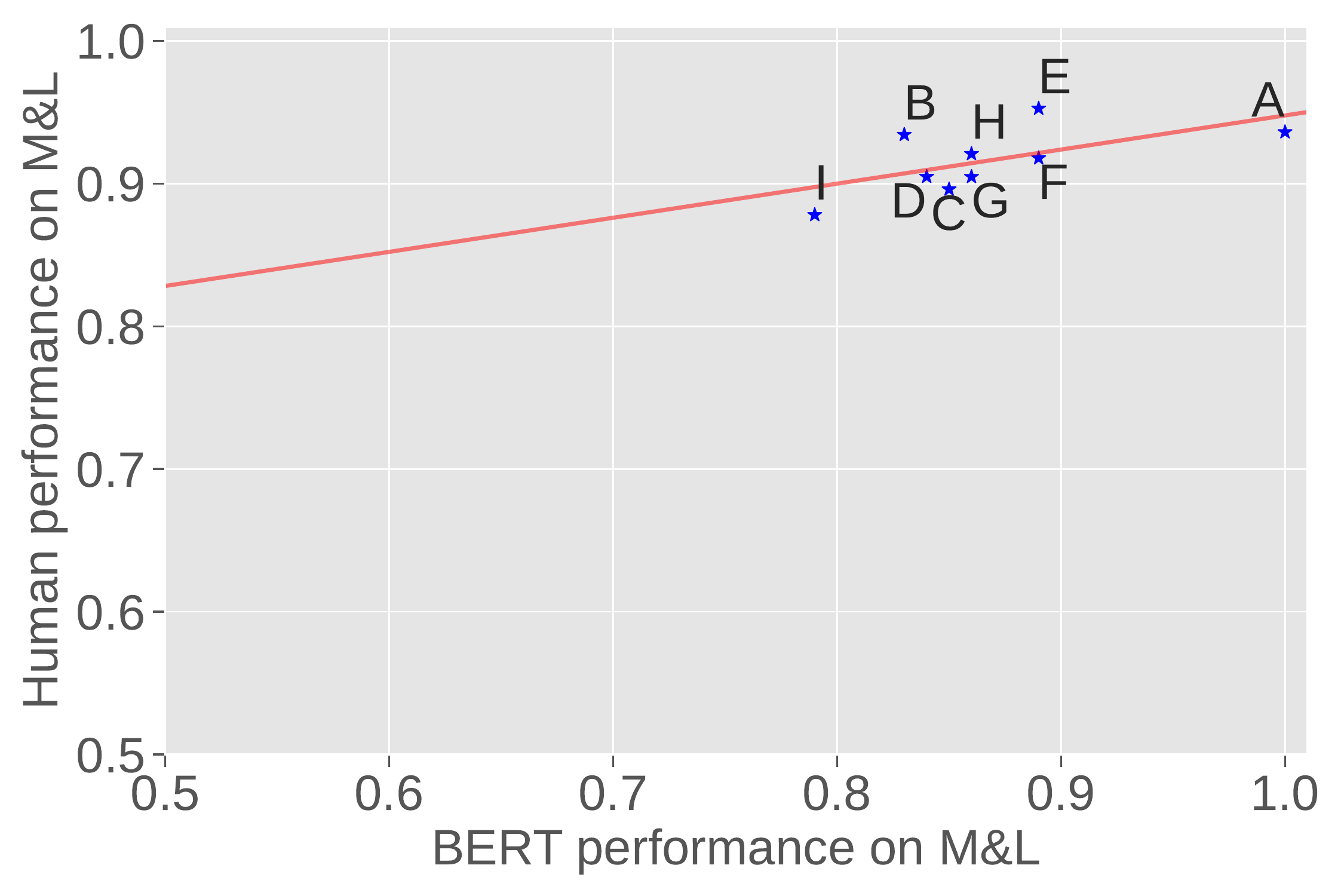}
        \subcaption{Human M\&L performance vs. BERT}
        \label{fig:ml-bert-humans}
    \end{subfigure}
    \caption{A comparison of human performance against BERT's performance on each of our structures.}
    \label{fig:comparison-human-BERT}
\end{figure*}

\paragraph{Collection of Human Judgements}

We collect our human data using the online click working platform Prolific.\footnote{\url{http://prolific.co}} We implemented a binary choice experiment in Psychopy \cite{peirce2007psychopy} hosted on Pavlovia\footnote{\url{https://pavlovia.org}} where participants were presented with a minimal pair, such as in \ref{admin_sings}, and asked which sentence was the most correct. In order to prevent habituation to our stimuli and task, we used 64\% of filler items. We recruited 300 participants to obtain 20 responses per item and kept the responses of 270 participants in our final data set. Their mean speed to judge one item was 6.9 seconds. See \cref{app:collection_hum_data} for more details about human data collection.

\section{Results}
We first discuss the human data, which we then compare with BERT's performance.

\subsection{Error Patterns for Humans}
In this analysis, we compare human accuracy on the nonce stimuli and on \citeposs{marvin-linzen-2018-targeted} sentences. In \cref{fig:raw-accuracies}, we break down the results by syntactic structure to observe whether the construction type affects the human judgments.
We notice a performance drop in all structures with nonce sentences, except for A where the apparent increase is not significant, as shown by the error bars.  Interestingly, the structures with an attractor (bolded in the \textit{x}-axis) are those for which the performance drop seems to be the highest. 
We also observe high performance drops in sentences where there is no attractor (B, G and I). 

Looking at \cref{tab:agreement-structures}, we can see that these structures are more complex than the structures where the effect of meaningfulness is low (A and E). Indeed, they contain either a complement (B), or a relative clause (G, I).
Surprisingly, we observe a similar pattern on meaningless sentences in (F) and (G): humans seem to be perturbed as much by the attractor within the object relative clause (F), as they are by the material in the main clause (G), if sentences are meaningless. This evidence in comprehension seems opposite to \citeposs {BOCK199299} claim that agreement production is only sensitive to information within the clause of the target. This evidence hints at the possibility of a difference in the mechanisms that support SVA in production and comprehension.

\begin{table}[h]
\centering
 \begin{tabular}{p{4cm}p{3cm}}
 \hline
 Metric & Correlation \\
 \hline\hline
 M\&L Accuracy & 0.61 \\
 \hline \hline
 Nonce Accuracy & 0.65 \\
 \hline \hline
 Accuracy Drop & 0.52 \\
 \hline \hline
 \end{tabular}
\caption{Coefficient of determination between BERT's and human performance on SVA, averaged across syntactic structures. The accuracy drop condition represents the difference between average performance on M\&L's stimuli and our nonce stimuli, as seen in \cref{fig:drop-bert-humans}.}
\label{tab:correlation}
\end{table}

\begin{figure}
    \begin{center} \includegraphics[width=\columnwidth]{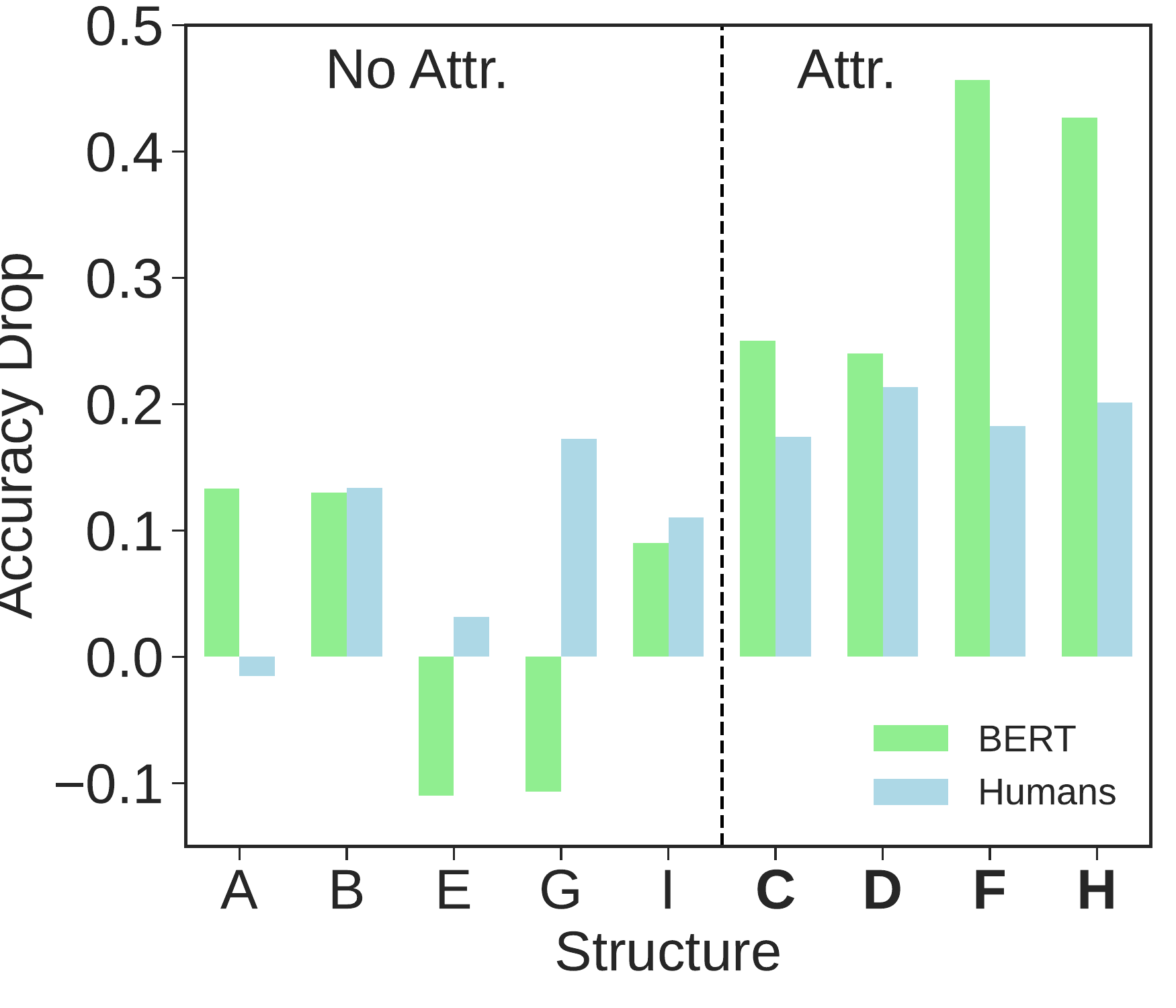}
    \caption{Performance drops between M\&L and nonce stimuli.}
    \label{fig:drop-bert-humans}
    \end{center}
\end{figure}

\subsection{How Similar are the Human and BERT's Error Patterns?}

In this analysis, we compare the performance achieved by BERT with human performance, on each of our stimuli types.
\cref{fig:comparison-human-BERT} displays the result obtained by humans against BERT's performance for each syntactic template, for both meaningful and meaningless sentences. Interestingly, there seems to be a fairly high alignment between the results for each syntactic construction, and for each source of stimuli. We display the $R^2$ correlation measurement of our fit in \cref{tab:correlation}. The latter confirms the observed alignment, as we obtain quite high correlations (0.61 for meaningful sentences and 0.65 for nonce sentences).

However, we observe that while the variation in performance obtained by humans across templates seems quite low, BERT's performance does seem to be more affected by the different structures. This is especially true in the case of nonce sentences, as seen in \cref{fig:nonce-bert-humans}. We also observe a difference in performance decrease in \cref{fig:drop-bert-humans}, as BERT's performance drops are overall higher in presence of an attractor compared to those of humans. On the other hand, BERT has a higher performance drop on (A) and humans on (G). This in turn could be explained by the fact that (G) is a hard sentence to process for humans, the target of the agreement being within an embedded relative clause, while BERT could rely on local context in this case as observed by \citet{lasri-etal-2022-bert}.

\section{Discussion}
\subsection{Lexicalization and Syntactic Generalization}
While  subject-verb agreement is sometimes considered as a purely syntactic phenomenon, our results show that actually humans also rely on semantics, which goes against the meaning independence hypothesis. Our results also show that BERT is also highly dependent on semantics, a finding in line with \citet{bernardy-lappin-2017-using}, who mention that ``\textit{deep neural networks require large vocabularies to form substantive lexical embeddings in order to learn structural patterns}''. This highlights the strong connection between the ability to process linguistic structure and the semantic content of sentences. 

\subsection{Structure Dependence}
Throughout this study, we observed that the performance of both humans and BERT were sensitive to the syntactic structure used in our items.
Humans clearly obtain lower performance on sentences that are more complex to process when they are meaningless, including but not limited to sentences presenting an attractor. This variation in performance seems to reflect variation in structure complexity, which upholds SD. On the other hand, BERT seems to be mostly sensitive to sentences with attractors. This evidence rather shows a violation of SD, as attractors are only related to the target by linear order, in line with evidence found by \citet{lasri-etal-2022-bert}. While human and BERT's results seem to correlate to a large extent, these divergences could reflect a difference in processing. For instance, SVA in sentence comprehension for humans could depend on having read the whole sentence, while BERT could rely more on local context for this task. Indeed, a fine-grained analysis performed in previous work showed BERT to be mostly sensitive to the replacement of linearly close tokens \cite{lasri-etal-2022-bert}. 

\section{Conclusion}
Throughout this work, we have shown that the ability to perform SVA is highly dependent on the syntactic construction when presented with meaningless sentences. The failures of humans seem to align well with those of BERT overall, as sentences with attractors tend to compromise meaning independence when processing the agreement relation. Despite these similarities, we further show that the performance drop is generally higher in BERT on meaningless sentences, and that humans are more perturbed by complex constructions without an attractor. This finding can in turn reflect differences in processing syntactic structure, either reflecting more reliance on local context for BERT, or a difference between agreement processing in production and comprehension, which could be the source of the partial mismatch in the observed error patterns. 


\section*{Ethics Statement}
The authors foresee no ethical concerns with the work presented in this paper.

\section*{Acknowledgements}
This work was funded in part by the French government under management of Agence Nationale de la
Recherche as part of the ``Investissements d’avenir''
program, reference ANR-19-P3IA-0001 (PRAIRIE
3IA Institute).

\bibliography{anthology, custom}


\appendix

\section{Vocabulary Selection}\label{selection_voc}
We used the vocabulary of \citet{lasri-etal-2022-bert} 
for the generation of our nonce items and filtered it. We selected a vocabulary of nouns and verbs that checked the following criteria:

\begin{enumerate}
\item We filter out tokens that are ambiguous, i.e. tokens which can either be a noun or a verb. We used Wordnet \cite{miller1995wordnet} implemented in the NLTK library \cite{bird2009natural} in python 3 to check whether a word was not classified as a noun and a verb by checking whether there was no synset in the other category. 
\item We filter out words using their relative frequency measured by using the python library wordfreq \citep{robyn_speer_2018_1443582}.\footnote{\url{https://pypi.org/project/wordfreq/}} We choose to filter too frequent words because some were ambiguous with another category (\textit{e.g.} in the noun vocabulary we can find \textit{good, well, one}). We decided to remove infrequent words to prevent that participants would not know their meaning. For example, this filtered out \textit{polynomial}, and \textit{consonant} from the noun vocabulary. 
\item We only keep words with a length ranging from 3 and 8 characters, in order to prevent big differences in size between the items produced by one template.
\item We make a subdivision between transitive and intransitive verbs in order to correctly fill the templates.
\end{enumerate}

\section{Collection of Human Judgments}\label{app:collection_hum_data}

\paragraph{Setup}
To collect our human judgements, we recruited participants on the working platform Prolific. Participants were redirected to the Pavlovia page hosting our experiment. First, they had to give their informed consent. Their data was processed in accordance to the European General Data Protection Regulation \citep{EUdataregulations2018}, and no sensitive data has been collected. After being informed, participants were shown brief instructions about the forced choice task. For each item, they were presented with two sentences, and asked to select the one that seemed more acceptable using the keyboard arrows. Each session started with three training items followed by feedback. When the training was finished, they were notified that the experiment started and that they would not receive feedback anymore. Each participant was presented with 100 items and thereafter received a message that the experiment was over. 

\paragraph{Number of Items and Participants} In total, the participants replied to 100 items: 64 fillers, and 36 experimental item. 18 were from the nonsense condition and 18 from the M\&L data set. As every condition features 9 different structures, 2 structures of each category where shown per participant. In order to collect 20 responses per item, we recruited 300 participants with 15 different versions of the online experiment, which can be found in the supplement material of this article.

\paragraph{Fillers} Our filler items where from \citet{ettinger-2020-bert}. They feature semantically appropriate and inappropriate completions. We also used filler items with correct and incorrect determiners among `a/an' depending on the following noun to feature syntax-oriented fillers as well. 

\paragraph{Selection of Participants} We only accepted participants with the United States nationality with English as a first language between the age 18 and 60 years old. We excluded participants that already contributed to another version of the experiment. 

\paragraph{Reward} Participants got rewarded 2.25£ for a participation that was estimated to take 15 minutes. This was estimated to be a suitable hourly pay by the Prolific platform. We rejected participants that performed the experiment in less than 5 minutes.

\paragraph{Filtering and Loss of Participants} In our final data set we have the contributions of 270 participants. Participants that performed at chance level (50~\% accuracy) were filtered out. Furthermore, we lost some data of participants that did not close the experiment correctly in Pavlovia.

\end{document}